\documentclass[10pt,twocolumn,letterpaper]{article}

\usepackage{lineno}
\usepackage{colortbl}
\usepackage{amsmath,amssymb,amsfonts}
\usepackage{algorithmic}
\usepackage{graphicx}
\usepackage{float}

\usepackage{cvpr}              % To produce the CAMERA-READY version
\definecolor{cvprblue}{rgb}{0.21,0.49,0.74}
\usepackage[pagebackref,breaklinks,colorlinks,allcolors=cvprblue]{hyperref}
\usepackage{multirow}
\usepackage{booktabs}
\usepackage{color}
\usepackage{colortbl}
\usepackage{graphicx}
\newcommand{\tinytit}[1]{\noindent\textbf{#1.}}

\def\paperID{6284} % *** Enter the Paper ID here
\def\confName{CVPR}
\def\confYear{2025}

\title{Evaluating Image Caption via Cycle-consistent Text-to-Image Generation}

\author{Tianyu Cui$^1$\thanks{Work done during the internship at AI Business, Alibaba Group.}   \ \ \ Jinbin Bai$^2$ \ \ \ Guo-Hua Wang$^2$ \ \ \ Qing-Guo Chen$^2$ \ \ \ Zhao Xu$^2$ \\ Weihua Luo$^2$ \ \ \ Kaifu Zhang$^2$ \ \ \ Ye Shi$^1$\thanks{Corresponding author, email: shiye@shanghaitech.edu.cn.} \\
$^1$ ShanghaiTech University \ \ \ $^2$ AI Business, Alibaba Group\\
{\tt\small \{cuity2022,shiye\}@shanghaitech.edu.cn} \\ {\tt\small \{baijinbin.bjb,  wangguohua,qingguo.cqg,changgong.xz,weihua.luowh,kaifu.zkf\}@alibaba-inc.com}\\
}

\begin{document}
\maketitle
\begin{abstract}
Evaluating image captions typically relies on reference captions, which are costly to obtain and exhibit significant diversity and subjectivity. While reference-free evaluation metrics have been proposed, most focus on cross-modal evaluation between captions and images. Recent research has revealed that the modality gap generally exists in the representation of contrastive learning-based multi-modal systems, undermining the reliability of cross-modality metrics like CLIPScore.
In this paper, we propose CAMScore, a cyclic reference-free automatic evaluation metric for image captioning models. 
To circumvent the aforementioned modality gap, CAMScore utilizes a text-to-image model to generate images from captions and subsequently evaluates these generated images against the original images.
Furthermore, to provide fine-grained information for a more comprehensive evaluation, we design a three-level evaluation framework for CAMScore that encompasses pixel-level, semantic-level, and objective-level perspectives.
Extensive experiment results across multiple benchmark datasets show that CAMScore achieves a superior correlation with human judgments compared to existing reference-based and reference-free metrics, demonstrating the effectiveness of the framework.

\end{abstract}    
\section{Introduction}
\label{sec:intro}

\begin{figure}[t]
    \centering
    \includegraphics[width=\linewidth]{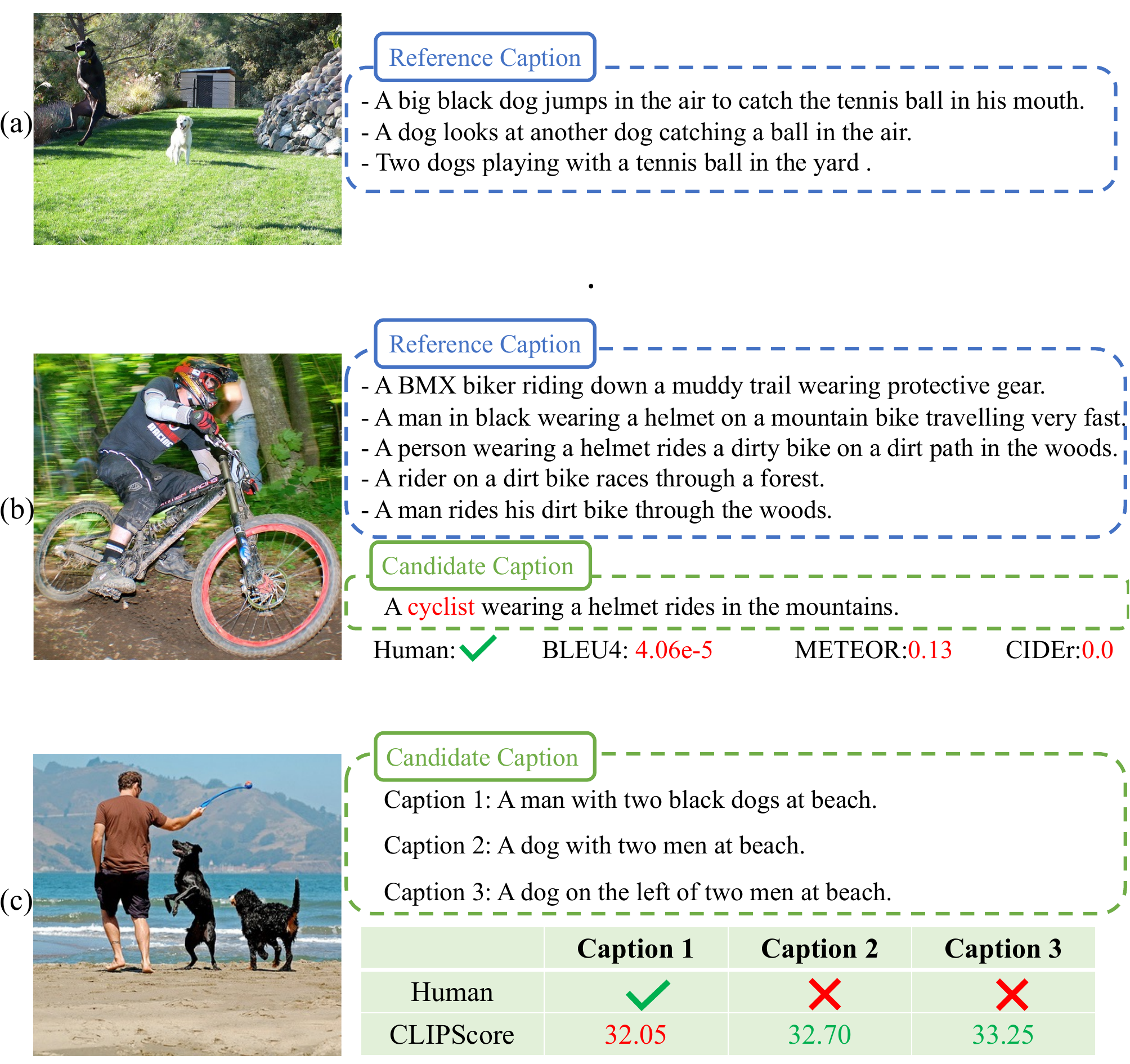}
    % \fbox{\rule{0pt}{3in} \rule{\linewidth}{0pt}} 
    \caption{(a) References fail to capture all information in the image, such as the color and position of the white dog. (b) Caption aligns with human judgment, but scores low on reference-based metrics. This discrepancy arises because different caption styles can lead to misalignment between reference-based metric scores and human judgments. (c) Modality gap in cross-modality evaluation can lead to confusion of attributes including numeracy and spatial relationships.}
    \label{fig:intro}
\end{figure}
High-quality image captions are vital for training a product-level vision foundation model. As a task to generate descriptive textual captions for given images, image captioning has garnered substantial attention from the research community in recent years \cite{xu2015show, anderson2018bottom, gurari2020captioning, hu2021question, bai2024meissonic}.
Furthermore, establishing an automatic evaluation metric that closely aligns with human judgment is crucial for advancing image captioning models effectively.
Previous research \cite{hessel2021clipscore} has demonstrated that reference-only evaluation metrics correlate poorly with human judgments. As \Cref{fig:intro} (a) shows, references alone often fail to fully capture the content of an image, leading to less reliability of reference-only evaluation image captioning metrics. Consequently, several approaches \cite{jiang2019tiger,lee2020vilbertscore,wang2021faier} have been proposed to integrate both references and images in the evaluation of image captions.

However, the diversity and subjectivity of reference captions pose significant challenges for evaluating image captioning performance \cite{lee2021umic}. Additionally, collecting reference captions can be resource-intensive, and even the availability of multiple human-authored captions per image often proves insufficient for comprehensive assessment \cite{hessel2021clipscore, lee2021umic}. As illustrated in \Cref{fig:intro} (b), various caption styles can lead to misalignment between reference-based metric scores and human judgments.

To address these limitations, reference-free evaluation metrics have been proposed, which directly utilize images instead of reference captions in the evaluation process.
Nonetheless, previous reference-free evaluation metrics predominantly focus on cross-modality evaluation between captions and images. Recent studies \cite{liang2022mind, shi2023towards} have revealed that the modality gap generally exists in the representation of contrastive learning-based multi-modal systems, undermining the reliability of cross-modality evaluation metrics like CLIPScore \cite{hessel2021clipscore}. 
As shown in \Cref{fig:intro} (c), CLIPScore, which relies solely on overall similarity, lacks fine-grained evaluation capabilities and may confuse attributes including numeracy and spatial relationships.

In this paper, we proposed CAMScore, a cyclic reference-free automatic evaluation metric for image captioning models. To circumvent the aforementioned modality gap, CAMScore utilizes a text-to-image model to generate images from captions and subsequently evaluates these generated images against the original images. By performing evaluations within the same image modality, CAMScore enables reference-free evaluation while addressing the modality gap inherent in existing cross-modality metrics, which exploits the cycle consistency across modalities.

Furthermore, to provide fine-grained information for a more comprehensive evaluation, we design a three-level evaluation framework for CAMScore. Specifically, this modular framework encompasses pixel-level, semantic-level, and objective-level evaluations, offering comprehensive evaluation from three distinct perspectives and thereby enhancing the robustness and accuracy of the metric.

To verify the effectiveness of CAMScore, we conduct extensive experiments across multiple image captioning benchmarks, aiming to assess the consistency of the proposed metric with human judgments. Specifically, we calculate the Kendall correlation coefficient on Flicker8k-Expert \cite{hodosh2013framing}, Flicker8k-CF \cite{hodosh2013framing} and COMPOSITE \cite{aditya2015images} datasets, and evaluate the pairwise ranking accuracy on PASCAL-50S \cite{vedantam2015cider} dataset. Our experimental results demonstrate that CAMScore achieves a superior correlation with human judgments compared to existing reference-based and reference-free metrics.

In summary, the main contributions of this paper are three-fold:
\begin{itemize}
    \item We proposed CAMScore, a novel cyclic reference-free evaluation metric for image captioning. By leveraging cycle consistency, CAMScore enables reference-free evaluation while addressing the modality gap inherent in existing cross-modality metrics, thereby facilitating evaluation within the same image modality.
    \item  We designed a three-level evaluation framework, which provides a modular framework for fine-grained evaluation of image captions from the pixel level, semantic level, and object level perspectives.
    \item Extensive experiments demonstrate that CAMScore achieves strong correlation with human judgments across various benchmarks including Flicker8k-Expert \cite{hodosh2013framing}, Flicker8k-CF \cite{hodosh2013framing}, COMPOSITE \cite{aditya2015images} and PASCAL-50S \cite{vedantam2015cider} datasets, showing the effectiveness of the proposed metric.
\end{itemize}

% \clearpage
\section{Related Work}
\label{sec:related}
\subsection{Image captioning evaluation metrics}
According to whether involving image and reference caption, image captioning evaluation metrics can be roughly divided into three categories: reference-only evaluation metrics, reference + image evaluation metrics, and reference-free evaluation metrics.\\

\tinytit{Reference-only evaluation metrics} This kind of metric \cite{papineni2002bleu,lin2004rouge,banerjee2005meteor,vedantam2015cider,anderson2016spice,zhao2019moverscore,zhang2019bertscore,yi2020improving,devlin2019bert} focus solely on comparing the reference caption with the candidate caption, primarily relying on $n$-grams or scene graphs. BLEU \cite{papineni2002bleu} calculates the precision between candidate and reference captions, ROUGE \cite{lin2004rouge} measures the recall of the longest common subsequence, and METEOR \cite{banerjee2005meteor} computes a version of alignment on unigram-level. CIDEr \cite{vedantam2015cider} employs tf-idf weighting for each n-gram, SPICE \cite{anderson2016spice} compares semantic propositional content using the predicted scene graph. BERTSCORE \cite{zhang2019bertscore} and its improved version \cite{yi2020improving} leverages learned embeddings from a pre-trained language model BERT \cite{devlin2019bert} to represent and measure semantic similarities. \\
% CLAIR \cite{chan2023clair} leverages the zero-shot language modeling capabilities of large language models to evaluate candidate captions. MoverScore \cite{zhao2019moverscore} combines contextualized representations of the candidate and reference texts and a distance between these representations.

\tinytit{Reference + image evaluation metrics}
References alone cannot fully capture the content of an image, making reference-only evaluation metrics for image captioning less reliable. To address this limitation, some approaches \cite{jiang2019tiger,lee2020vilbertscore,wang2021faier} integrate both reference captions and images to evaluate image captions. TIGEr \cite{jiang2019tiger} utilizing text-to-image grounding results based on a pre-trained SCAN model. ViLBERTScore \cite{lee2020vilbertscore} generates image-conditioned embeddings for each token in candidate and reference captions using ViLBERT \cite{lu2019vilbert}.  MID \cite{kim2022mutual} uses
CLIP visual-textual features to compute negative Gaussian cross-mutual information, resulting in a more effective evaluation metric.\\
% Although the reference is supplemented with the image for evaluation, these methods are still not applicable in the absence of the reference. Polos \cite{wada2024polos} is trained by directly regressing the human evaluation and computing scores from multimodal inputs. FAIEr \cite{wang2021faier} is a learning-based metric that combines the scene graphs of the image and reference captions as a union scene graph, which is compared with the scene graph of candidate captions. COSMic \cite{inan2021cosmic} is a coherence-aware embedding-based generation metric that correlates with human judgments.
%虽然使用图像信息弥补了文字，但是在没有reference的情况下这些方法仍然都不适用

\tinytit{Reference-free evaluation metrics}
% data quality, acquiring a set of reference captions can be expensive to collect, 
In the reference + image evaluation metrics, although the image supplements the reference for evaluation, these methods remain inapplicable in the absence of reference. Therefore, reference-free metrics \cite{lee2021umic,hessel2021clipscore,hu2023infometic,lee2024fleur} have been proposed to evaluate candidate captions without references. UMIC \cite{lee2021umic} fine-tunes a pre-trained UNITER \cite{chen2020uniter} via contrastive learning to compute the score of the caption. CLIPScore \cite{hessel2021clipscore} relies on the image-text similarity from the CLIP model \cite{radford2021learning}. Based on CLIPScore, PAC-S\cite{sarto2023positive} further fine-tunes CLIP with positive-augmented contrastive learning and evaluates captions in the same manner as CLIPScore. In addition to metrics based on CLIP, there are also methods based on the large multimodal model (LMM).
FLEUR \cite{lee2024fleur} leverages LMM to introduce explainability into image captioning evaluation.
Previous methods predominantly focus on cross-modality evaluation between captions and images, thereby failing to address the inherent modality gap.
Different from previous approaches, we proposed a novel cyclic evaluation metric, CAMScore, which leverages cycle consistency to perform evaluations within the same image modality, thereby circumventing the modality gap.
% InfoMetIC \cite{hu2023infometic} builds on top of pre-trained VLP models to measure fine-grained similarities between images and captions.
\subsection{Cycle consistency}
Our work is influenced by cycle consistency, a method to bridge one modality or domain to the other. Cycle consistency has been widely applied across various tasks, including unpaired image-to-image translation \cite{zhu2017unpaired,hoffman2018cycada}, depth estimation \cite{godard2017unsupervised}, visual object tracking \cite{wang2019unsupervised}, visual question answering \cite{shah2019cycle}, video-text retrieval \cite{bai2022lat} and image captioning \cite{guo2019mscap}. 
Unlike previous approaches that focus primarily on consistency within a single modality or task, our approach jointly forms a cyclic relationship between two modalities (text and image), and two tasks (image captioning and text-to-image task). 
To the best of our knowledge, cycle consistency between these two tasks has been rarely addressed in the literature. In this work, we leverage cycle consistency for image captioning evaluation, enabling the evaluation reference-free and circumventing the modality gap in existing cross-modality evaluation metrics.

\begin{figure}[t]
  \centering
  \includegraphics[width=\linewidth]{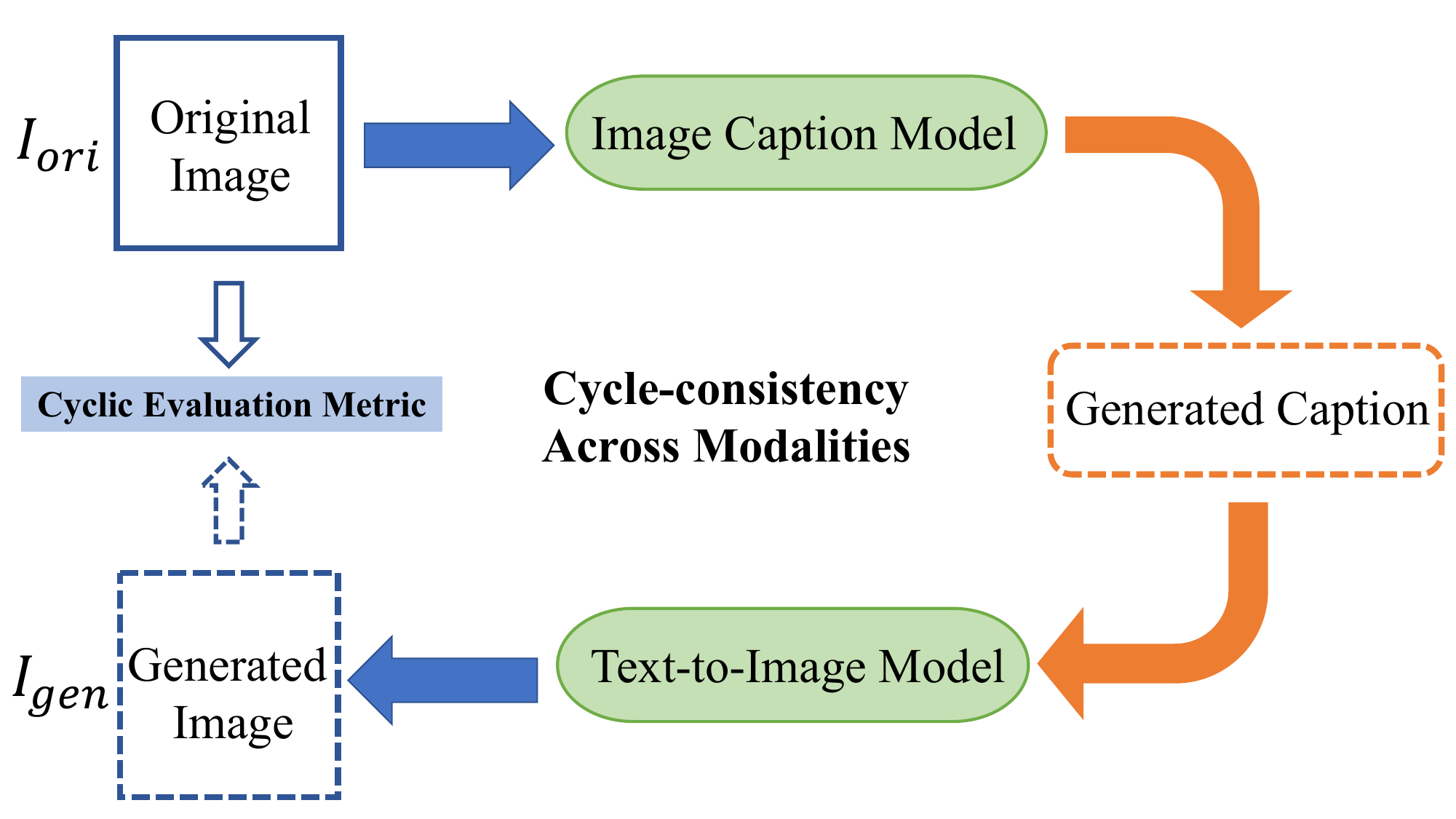}

   \caption{Overview of our proposed framework.}
   \label{fig:CAM}
\end{figure}

\begin{figure*}[t]
    \centering
    \includegraphics[width=\linewidth]{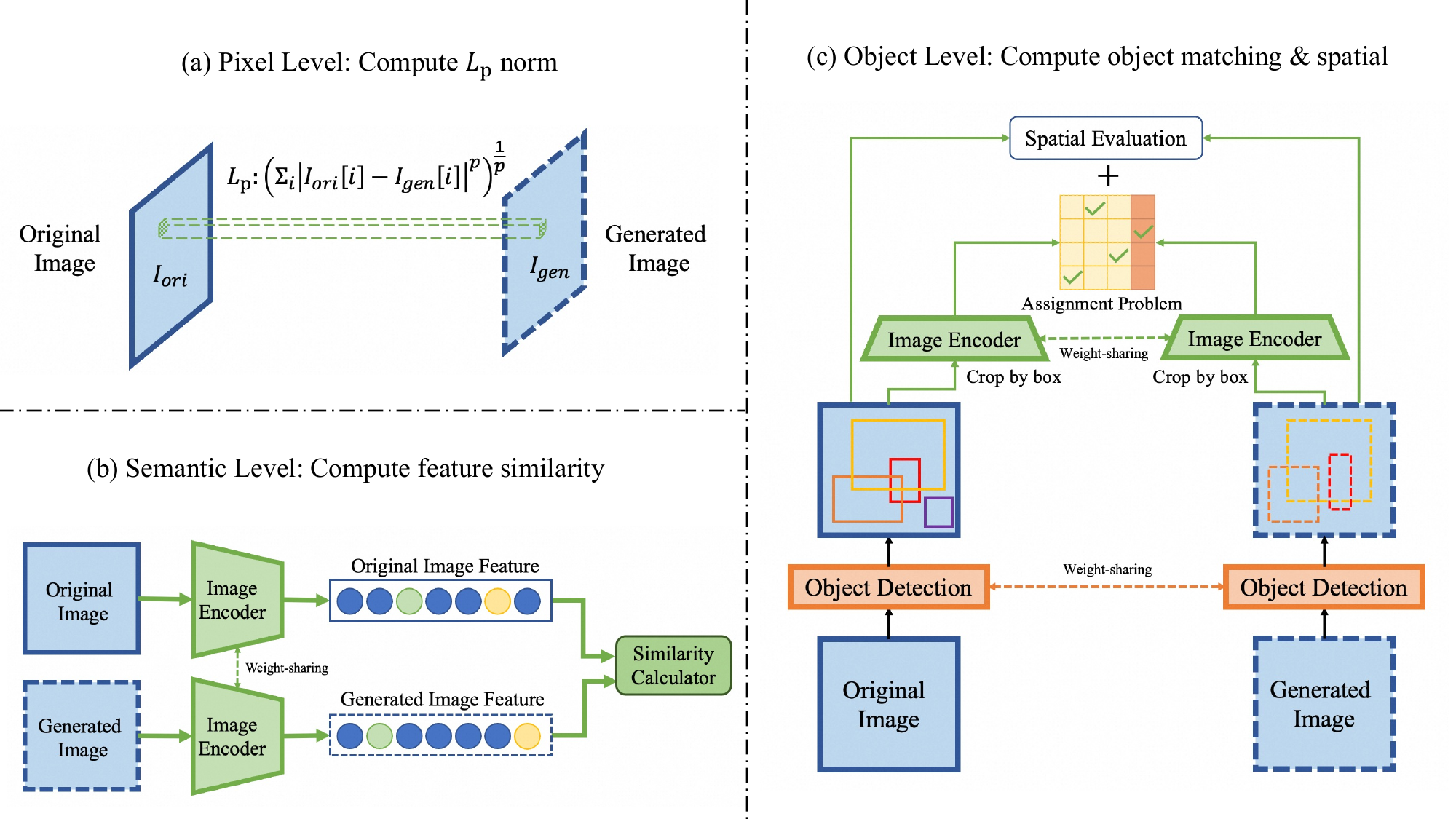}
    \caption{Illustration of our proposed evaluation metrics: (a) Calculate the pixel-by-pixel differences as pixel-level evaluation, (b) Calculate the similarity between the features of the original image and the generated image as semantic-level evaluation, (c) Detection-based object-level evaluation, taking into account both object matching and spatial relationship.}
    \label{fig:pipline}
    % \vspace{-.3cm}
\end{figure*}
\section{Cycle-consistent Evaluation Framework}
In this section, we delve into the details of our proposed cycle-consistent evaluation framework.
As illustrated in \Cref{fig:CAM}, CAMScore evaluates image caption models by employing a frozen text-to-image model to generate images $I_{gen}$ from the caption. These generated images are then compared against the original image $I_{ori}$ across three distinct perspectives. 
By performing evaluations within the same image modality, CAMScore enables reference-free evaluation while addressing the modality gap inherent in existing cross-modality metrics, which exploits the cycle consistency across modalities.
% By exploiting the cycle consistency across modalities, our method effectively avoids the modality gap existing in cross-modal measurements.

% We denote the Orignal Image as $I_{ori}$, then for the cycle-consistence across modalities, we got
% \begin{equation}
%     T_{gen} = G(I_{ori}),  I_{gen} = F(T_{gen}) = F(G(I_{ori})),
% \end{equation}
%  where $T_{gen}$ is the Generated Caption of Orignal Image $I_{ori}$, and $I_{gen}$ denote the Generated Image. $G: I \to T$ (Vision Large Language Model), $F: T \to I$ (Text-to -Image Model)
% We calculate the score between Original Image $I_{ori}$ and the Generated Image $I_{gen}$ using the Cyclic Evaluation Metric $D(I_{ori}, I_{gen})$.Thanks to cycle consistency, Cycle-GAN \cite{zhu2017unpaired} is capable of learning image-to-image translation in the absence of paired training data. inspired by this, 

\subsection{Image caption generation module}
For a given image, the image captioning model generates a descriptive text caption $T_{gen}$. The process can be formulated as follows:
\begin{equation}
    T_{gen} = G(I_{ori}),
\end{equation}
where $G(\cdot)$ and $T_{gen}$ denote the image caption model and the generated caption, respectively.
% A high-quality image caption should include the following descriptions:
% 1. Attribute binding (including color, shape, and texture),
% 2. Object relationship (including spatial relationship and non-spatial relationship)
% 3. Numeracy
% 4. Complex compositions. (e.g.: scissors cutting paper)
% 5. Image style

% \textit{
% Describe the image, the foreground and the background. Clarify positional information, colors, counts of objects, and other visual aspects and features. Make sure to include as much detail as possible. Make sure to describe the spatial relationships seen in the image. You can use words like left/right, above/below, front/behind, far/near/adjacent, inside/outside. Make sure to include object interactions like "a table is in front of the kitchen pot" and "there are baskets on the table". Also describe relative sizes of objects seen in the image. Make sure to include counts of prominent objects in the image, especially when there are humans in the image.  Be factual in your description, capturing the content, and style of the image. Describe the image in a short but descriptive manner within 256 tokens. Start with the words ‘This image displays:’}

\subsection{Text-to-image generation module}
We leverage the pre-trained text-to-image model to generate new images based on the image caption, transforming the caption from text modality to image modality. Specifically, the process of generating the image from the caption can be formulated as:
\begin{equation}
    I_{gen} = F(T_{gen}) = F(G(I_{ori})),
\end{equation}
where $F$ denotes the text-to-image model.
Our framework evaluates the generated image with the original image through the cyclic evaluation metric, eliminating the need for reference captions. Moreover, the cycle consistency in our approach is cross-modality, which transforms captions into images and compares them within the same image modality to mitigate the impact of the modality gap.

\subsection{Cyclic evaluation metric}
After we obtain the generated image based on the caption through the text-to-image model, we evaluate the generated image against the original image through the cyclic evaluation metric. Specifically, we elaborate a three-level evaluation framework $D(I_{ori}, I_{gen})$ in the following parts:\\

\tinytit{Pixel level} The pixel-level evaluation directly reflects the difference between difference between the generated image and the original image. In specific, we calculate the $L_p$ norm (Minkowski Distance) between the original image $I_{ori}$ and the generated image $I_{gen}$ by pixel level, as shown in (a) of \Cref{fig:pipline}. The calculation process can be formulated as follows:
\begin{equation}
    \mathcal{L}_{pix} = \left(\sum_i | I_{ori}[i] - I_{gen}[i] |^p \right)^{\frac{1}{p}}.
    \label{equ:pix}
\end{equation}
\Cref{equ:pix} enables a quantitative assessment of the pixel-wise discrepancies between the original and generated images, thereby providing a fine-grained evaluation within CAMScore.\\

\tinytit{Semantic level} 
To assess the overall semantic information of an image, we perform semantic-level evaluation utilizing a pre-trained image encoder as the image feature extraction module. This encoder takes an image as input to generate a feature vector that represents the semantic content of the input image. We then calculate the similarity between the features of the original image and the generated image. 
Specifically, we employ a pre-trained ViT-B16 \cite{radford2021learning} as an exemplary model for image feature extraction.  The ViT-B16 model is capable of generating rich and robust image representations, making it highly effective for various computer vision tasks. As for similarity calculation, we adopt cosine similarity as the metric:
\begin{equation}
    \mathcal{L}_{sem} = \frac{\langle \phi(I_{ori}), \phi(I_{gen})) \rangle }
    {\lVert \phi(I_{ori})\rVert_2  \cdot  \lVert\phi(I_{gen})\rVert_2},
\end{equation}
where $\phi(\cdot)$ denotes the image feature extraction module, $\lVert \cdot\rVert_2$ denotes the Euclidean norm of vectors and $\langle \cdot \; , \; \cdot \rangle$ denotes the inner product of vectors. This approach enables a holistic comparison of the semantic content between images, thereby enhancing the evaluation robustness of CAMScore.\\
% Then we got the semantic score between 0-1.

\tinytit{Object level} 
While semantic-level evaluation provides a holistic assessment of an entire image, it lacks the granularity required for detailed analysis. To achieve a more fine-grained evaluation, we introduce a detection-based object-level evaluation metric that yields detailed information. This approach involves applying an object detection model to both the original and generated images. For each detected object, the detection model outputs its corresponding bounding box.
Utilizing these bounding boxes, we crop the objects and extract their feature representations  $f_{ori}$ from the original image and $f_{gen}$ from the generated image using the image feature extraction module $\phi(\cdot)$:
\begin{equation}
    f_{ori} = \{\phi(B_{ori}^i(I_{ori}))\}_{i=1}^m,  f_{gen} = \{\phi(B_{gen}^j(I_{gen}))\}_{j=1}^n,
\end{equation}
where $\{B_{ori}^i\}_{i=1}^m$ and $\{B_{gen}^j\}_{j=1}^n$ denote the sets of $m$ and $n$ bounding boxes for the original images $I_{ori}$ and generated images $I_{gen}$, respectively. 
 % $:\{B_{ori}^1, B_{ori}^2, \ldots, B_{ori}^m \}$ $:\{B_{gen}^1, B_{gen}^2, \ldots, B_{gen}^n  \}$
Subsequently, we can compute the object-level similarity matrix between the original and generated images and derive the corresponding cost matrix:
\begin{equation}
    S [i,j]= \frac{\langle f_{ori}[i],f_{gen}[j] \rangle }
    {\lVert f_{ori}[i]\rVert_2  \cdot  \lVert f_{gen}[j]\rVert_2},
\end{equation}
\begin{equation}
    \begin{split}
        \text{Cost}[i,j] &= 1 - \frac{\langle f_{ori}[i],f_{gen}[j] \rangle }
    {\lVert f_{ori}[i]\rVert_2  \cdot  \lVert f_{gen}[j]\rVert_2}.
    \end{split}
\end{equation}

% \begin{equation}
%     S_{cos} [i,j]= \frac{\langle \phi(B_{ori}^i(I_{ori})),\phi(B_{gen}^j(I_{gen})) \rangle }
%     {\lVert \phi(B_{ori}^i(I_{ori}))\rVert_2  \cdot  \lVert\phi(B_{gen}^j(I_{gen}))\rVert_2},
% \end{equation}
% \begin{equation}
%     \begin{split}
%         Cost[i,j] &= 1 - S_{cos}[i,j] \\
%     &= 1 - \frac{<\phi(B_{ori}^i(I_{ori})), \phi(B_{gen}^j(I_{gen})))>}{||\phi(B_{ori}^i(I_{ori}))|| \cdot  ||\phi(B_{gen}^j(I_{gen}))||}
%     \end{split}
% \end{equation}
To penalize discrepancies in object counts between the original and generated images, when $m > n$, we pad the cost matrix to an  $m \times m$  matrix with elements set to $1$ (the maximum cost), thereby imposing a numerical penalty for missing objects.
We model the cost matrix as a classical assignment problem and employ the Hungarian (Kuhn-Munkres) algorithm \cite{kuhn1955hungarian, munkres1957algorithms} to determine the optimal object matching and compute the minimum matching cost:
\begin{equation}
    \begin{split}
    \mathcal{L}_{obj} = & \min_{X} \sum_i \sum_j \text{Cost}[i,j] X[i,j]\\
     s.t. \quad &\sum_{i=1}^m X[i,j] = 1, \quad j = 1,2 \dots n,\\
     &\sum_{j=1}^n X[i,j] = 1, \quad i = 1,2, \dots m,\\
     &X[i,j] \in \{0,1\}.
    \end{split}
\end{equation}

% \begin{equation}
%     \begin{split}
%         \mathcal{L}_{obj} &=  \min Cost(\phi(B_{ori}^i(I_{ori})), \phi(B_{gen}^j(I_{gen})))\\
%         &= \Sigma Cost[rowls,cols]
%     \end{split}
% \end{equation}

% \begin{equation}
%     [rows,cols] = \mathop{\arg\min}_{i,j} Cost(\phi(B_{ori}^i(I_{ori})), \phi(B_{gen}^j(I_{gen})))
% \end{equation}
% \begin{equation}
%     \begin{split}
%         \mathcal{L}_{obj} &=  \min Cost(\phi(B_{ori}^i(I_{ori})), \phi(B_{gen}^j(I_{gen})))\\
%         &= \Sigma Cost[rowls,cols]
%     \end{split}
% \end{equation}

However, the matching cost primarily captures the presence and semantic similarity of objects, while neglecting their spatial information. To address this limitation, we incorporate a 3D-spatial relationship evaluation for the matching object pairs. We leverage the object detection algorithms to obtain 2D positional information and utilize depth estimation to acquire depth information, thereby enabling 3D-spatial relationship evaluation between matched object pairs.
Specifically, for each box $B:(x_1,y_1,x_2,y_2)$, we calculate its relative position by normalizing it with respect to the size of the image and subsequently compute the Complete-IoU (CIoU) loss \cite{zheng2021enhancing}. CIoU loss takes into account the Intersection over Union (IoU), normalized central point distance, and the consistency of aspect ratio:
\begin{equation}
    \mathcal{L}_{CIoU} = 1 - \text{IoU} + \frac{d^2}{c^2} + \alpha v,
\end{equation}
where $c$ is the diagonal length of the smallest enclosing box covering two boxes, and $d$ is the distance of central points of two boxes,  $\alpha$ is the trade-off parameter, $v$ measures the consistency of the aspect ratios between the predicted and ground truth boxes.
% , typically defined as:
% \begin{equation}
%     v = \frac{4}{\pi^2} \left( \arctan\frac{w^{gt}}{h^{gt}} - \arctan\frac{w}{h} \right)^2
% \end{equation}
Additionally, we incorporate the scale-invariant error \cite{eigen2014depth} to further refine the spatial relationship evaluation:
\begin{equation}
    \mathcal{L}_{dep}  =  \frac1n \sum_i d_i^2 - \frac1{n^2} \left(\sum_i d_i \right)^2,
\end{equation}
where $d_i = \log (I_{gen}[I]) - \log (I_{ori}[I])$ is the difference between the prediction and ground truth at pixel $i$, and $n$ denotes the total number of pixels.
The scale-invariant loss function measures the error magnitude by focusing only on the relative depth of each value set without considering the scale difference between the ground truth and the prediction.
This combination of metrics enables a more comprehensive assessment of both the geometric alignment and depth consistency between matched objects. Therefore, our evaluation framework achieves enhanced robustness by effectively capturing semantic and spatial nuances in object relations.
Subsequently, following the approach of \cite{wada2024polos}, we employ a multilayer perceptron (MLP) to integrate the scores and compute the final score:
\begin{equation}
    \text{CAMScore} = \text{MLP}(\mathcal{L}_{pix}, \mathcal{L}_{sem}, \mathcal{L}_{obj}, \mathcal{L}_{CIoU}, \mathcal{L}_{dep})
\end{equation}
For the loss function, we adopted the Mean Squared Error (MSE), a conventional choice in regression tasks due to its proven efficacy in quantifying the discrepancy between predicted values and ground truth human judgments. Additional details can be found in the Appendix.

\section{Experiments}

\begin{table}[t]
\large
\centering
\setlength{\tabcolsep}{.2em}
\resizebox{\linewidth}{!}{
\renewcommand\arraystretch{1.1}
\begin{tabular}{lc cc c cc}
\toprule
\multirow{2}*{\textbf{Metric}} & & \multicolumn{2}{c}{\textbf{Flickr8k-Expert}} & & \multicolumn{2}{c}{\textbf{Flickr8k-CF}} \\
\cmidrule{3-4} \cmidrule{6-7}
 & & Kendall $\tau_b$ & Kendall $\tau_c$  & & Kendall $\tau_b$ & Kendall $\tau_c$ \\
\midrule
 BLEU-1~\cite{papineni2002bleu} & & 32.2 & 32.3 & & 17.9 & 9.3 \\
 BLEU-4~\cite{papineni2002bleu} & & 30.6 & 30.8 & & 16.9 & 8.7 \\
 ROUGE~\cite{lin2004rouge} & & 32.1 & 32.3 & & 19.9 & 10.3 \\
 METEOR~\cite{banerjee2005meteor} & & 41.5 & 41.8 & & 22.2 & 11.5 \\
 CIDEr~\cite{vedantam2015cider} & & 43.6 & 43.9 & & 24.6 & 12.7 \\
 SPICE~\cite{anderson2016spice} & & 51.7 & 44.9 & & 24.4 & 12.0 \\
 BERT-S~\cite{zhang2019bertscore} & & - & 39.2 & & 22.8 & - \\
 BERT-S++~\cite{yi2020improving} & & - & 46.7 & & - & - \\
 TIGEr~\cite{jiang2019tiger} & & - & 49.3 & & - & - \\
 ViLBERTScore~\cite{lee2020vilbertscore} & & - & 50.1 & & - & - \\
 MID~\cite{kim2022mutual} && - & \underline{54.9} && \underline{37.3} & - \\
\midrule
 UMIC~\cite{lee2021umic} & & - & 46.8 & & - & - \\
 CLIP-S~\cite{hessel2021clipscore} &&  51.1 & 51.2 & & 34.4 & 17.7 \\
 PAC-S~\cite{sarto2023positive} & & \underline{53.9} & 54.3 & & 36.0 & \underline{18.6} \\

 \textbf{CAMScore} & & \textbf{54.8} & \textbf{55.6} & & \textbf{37.5} & \textbf{19.3}  \\
\bottomrule
\end{tabular}
}
 \vspace{-0.1cm}
\caption{The Kendall correlation coefficient $\tau_b$ and $\tau_c$ between
human judgments and various automatic metrics on Flickr8k-Expert and Flickr8k-CF~\cite{hodosh2013framing} datasets. Bold font indicates the highest recorded value overall, while underlining indicates the second-highest value. The first part is for reference-based methods, and the second part is for reference-free methods.}
\label{tab:flickr}
\vspace{-0.35cm}
\end{table}

\subsection{Evaluation datasets} 
Human correlation and accuracy represent critical measures for evaluating image captioning metrics. To assess the caption-level correlation between automatic metrics and human judgments, we utilize the Flicker8k \cite{hodosh2013framing} and Composite dataset \cite{aditya2015images},  wherein human annotators provide evaluations for each candidate caption.  Following previous works \cite{hessel2021clipscore,lee2021umic,sarto2023positive}, we compute Kendall’s correlation coefficient $\tau_b$ and $\tau_c$ to measure the alignment between human judgments and metric scores. Kendall’s correlation coefficient is a statistic used to measure the ordinal association between two measurements. For datasets of a different nature, like Pascal-50S \cite{vedantam2015cider}, where annotators are tasked with selecting the superior caption from pairs of candidate captions, we assess performance using metric pairwise ranking accuracy. \\

\begin{table}[t]
\small
\centering
\setlength{\tabcolsep}{.55em}
\resizebox{0.8\linewidth}{!}{
\renewcommand\arraystretch{1}
\begin{tabular}{lc cc}
\toprule
\multirow{2}*{\textbf{Metric}} & & \multicolumn{2}{c}{\textbf{Composite}} \\
\cmidrule{3-4}
& & Kendall $\tau_b$ & Kendall $\tau_c$ \\
\midrule
BLEU-1~\cite{papineni2002bleu} & & 29.0 & 31.3 \\
BLEU-4~\cite{papineni2002bleu} & & 28.3 & 30.6 \\
ROUGE~\cite{lin2004rouge} & & 30.0 & 32.4 \\
METEOR~\cite{banerjee2005meteor} & & 36.0 & 38.9 \\
CIDEr~\cite{vedantam2015cider} & & 34.9 & 37.7 \\
SPICE~\cite{anderson2016spice} & & 38.8 & 40.3 \\
BERT-S~\cite{zhang2019bertscore} & & - & 30.1 \\
BERT-S++~\cite{yi2020improving} & & - & 44.9 \\
TIGEr~\cite{jiang2019tiger} & & - & 45.4 \\
ViLBERTScore~\cite{lee2020vilbertscore} & & - & 52.4 \\
MID~\cite{kim2022mutual} && - & 55.7 \\
\midrule
UMIC~\cite{lee2021umic} && - & \underline{56.1} \\
CLIP-S~\cite{hessel2021clipscore} & & 49.8 & 53.8 \\
PAC-S~\cite{sarto2023positive} & & \underline{51.5} & 55.7 \\
\textbf{CAMScore} & & \textbf{53.4} & \textbf{57.5} \\
\bottomrule
\end{tabular}
}
\vspace{-0.1cm}
\caption{The Kendall correlation coefficient $\tau_b$ and $\tau_c$ between
human judgments and various automatic metrics on the Composite dataset~\cite{aditya2015images}.  Bold font indicates the highest recorded value overall, while underlining indicates the second-highest value.}
 % The symbol ‘–’ indicates non-executable code or unavailable data.
\label{tab:composite}
\vspace{-0.35cm}
\end{table}

\tinytit{Flickr8k-Expert} \cite{hodosh2013framing} Flickr8k-Expert dataset contains 17k expert annotations for image-caption pairs, encompassing a total of 5,664 distinct images. Each image-caption pair is evaluated by three expert annotators with scores ranging from 1 (irrelevant) to 4 (perfect match).\\

\tinytit{Flickr8k-CF} \cite{hodosh2013framing} Flickr8k-CF consists of 145k binary quality judgments, collected from CrowdFlower, covering 48k image-caption pairs that contain 1k unique images. Each pair involves three annotators to determine whether the image-caption pair matches, categorized as either “yes” or “no”. The score for each pair is the proportion of “yes” responses.\\

\tinytit{Composite} \cite{aditya2015images} Composite dataset comprises human judgments of 12,000 image-caption pairs, incorporating 3,995 images from MSCOCO (2,007 images) \cite{lin2014microsoft}, Flickr30K (991 images) \cite{young2014image}, and Flickr8k (997 images) \cite{hodosh2013framing}. Human evaluators were asked to rate each image-caption pair and assign a score on a scale of 1 to 5 to estimate how well the caption is aligned with the associated image.\\

\tinytit{Pascal-50S} \cite{vedantam2015cider} Pascal-50S includes 4,000 caption pairs associated with 1,000 images, along with a label indicating which of the two captions is deemed correct by 48 annotators. Each image is linked to approximately 50 reference captions. In Pascal-50S, caption pairs are categorized based on the composition of the two captions: HC denotes two correct human-written captions; HI indicates one correct and one incorrect human-written caption; HM represents one human-written caption and one machine-generated caption; and MM denotes two machine-generated captions.

\subsection{Baselines}
\label{subsec:baselines}

\begin{table}[t]
\normalsize
\centering
\setlength{\tabcolsep}{.55em}
\resizebox{\linewidth}{!}{
\renewcommand\arraystretch{1.05}
\begin{tabular}{lc cccc c c}
\toprule
\textbf{Metric} & & HC & HI & HM & MM & & Mean \\
\midrule
BLEU-1~\cite{papineni2002bleu} & & 64.6 & 95.2 & 91.2 & 60.7 & & 77.9 \\
BLEU-4~\cite{papineni2002bleu} & & 60.3 & 93.1 & 85.7 & 57.0 & & 74.0 \\
ROUGE~\cite{lin2004rouge} & & 63.9 & 95.0 & 92.3 & 60.9 & & 78.0 \\
METEOR~\cite{banerjee2005meteor} & & 66.0 & 97.7 & 94.0 & 66.6 & & 81.1 \\
CIDEr~\cite{vedantam2015cider} & & 66.5 & 97.9 & 90.7 & 65.2 & & 80.1 \\
SPICE~\cite{anderson2016spice} & & 63.6 & 96.3 & 86.7 & 68.3 && 78.7 \\
BERT-S~\cite{zhang2019bertscore} & & 65.4 & 96.2 & 93.3 & 61.4 & & 79.1 \\
BERT-S++~\cite{yi2020improving} & & 65.4 & 98.1 & 96.4 & 60.3 & & 80.1 \\
TIGEr~\cite{jiang2019tiger} & & 56.0 & \textbf{99.8} & 92.8 & 74.2 & & 80.7 \\
ViLBERTScore~\cite{lee2020vilbertscore} & & 49.9 & 99.6 & 93.1 & 75.8 & & 79.6\\
MID~\cite{kim2022mutual} & & 67.0 & \underline{99.7}& \underline{97.4} & 76.8 & & \underline{85.2} \\  
\midrule
UMIC~\cite{lee2021umic}  & & 66.1 & \textbf{99.8} & \textbf{98.1} &76.2 & &   85.1\\
CLIP-S~\cite{hessel2021clipscore} & & 55.9 & 99.3 & 96.5 & 72.0 & & 80.9 \\
PAC-S~\cite{sarto2023positive} && 60.6 & 99.3 & 96.9 &72.9& & 82.4\\
\textbf{CAMScore} & & \textbf{68.8} & 99.6 & \underline{97.4} & \textbf{77.4} & & \textbf{85.8} \\

\bottomrule
\end{tabular}
\vspace{-0.1cm}
\label{tab:pascal}
}
% FAIEr$^\dagger$~\cite{wang2021faier} & & 59.7 & {\textbf{99.9}} & 92.7 & 73.4 & & 81.4 \\
\caption{Caption pairwise ranking accuracy results on the Pascal-50S dataset~\cite{vedantam2015cider}.
obtained by averaging the scores over five random draws of reference captions (except for reference-free metrics). Bold font indicates the highest recorded value overall, while underlining indicates the second-highest value.}
\vspace{-0.35cm}
\end{table}
All baseline metrics have been briefly introduced in \Cref{sec:related}.
Specifically, we adopt BLEU \cite{papineni2002bleu}, ROUGE \cite{lin2004rouge}, METEOR \cite{banerjee2005meteor}, CIDEr \cite{vedantam2015cider}, SPICE \cite{anderson2016spice}, BERTSCORE \cite{zhang2019bertscore}, TIGEr \cite{jiang2019tiger}, ViLBERTScore \cite{lee2020vilbertscore}, and MID \cite{kim2022mutual} as baseline metrics due to their reliance on references for evaluating image captioning tasks. Additionally, we incorporate UMIC \cite{lee2021umic}, CLIPScore \cite{hessel2021clipscore}, and PAC-S \cite{sarto2023positive} as baseline metrics because they are reference-free evaluation metrics for image captioning.

\begin{table*}[!ht]
\normalsize
\centering
\setlength{\tabcolsep}{.55em}
\resizebox{\linewidth}{!}{
\renewcommand\arraystretch{1.2}
\begin{tabular}{lc cc c cc c cc c ccccc}
\toprule
\multirow{2}*{\textbf{Metric}} & & \multicolumn{2}{c}{\textbf{Flickr8k-Expert}~\cite{hodosh2013framing}} & &\multicolumn{2}{c}{\textbf{Flickr8k-CF}~\cite{hodosh2013framing}} & & \multicolumn{2}{c}{\textbf{Composite}~\cite{aditya2015images}} & & \multicolumn{5}{c}{\textbf{Pascal-50S}~\cite{vedantam2015cider}} \\ 
\cmidrule{3-4} \cmidrule{6-7} \cmidrule{9-10} \cmidrule{12-16}
& & Kendall $\tau_b$ & Kendall $\tau_c$ & & Kendall $\tau_b$ & Kendall $\tau_c$ & & Kendall $\tau_b$ & Kendall $\tau_c$ & & HC & HI & HM & MM & Mean \\
\midrule
w/o pixel && 53.9 (-0.9) & 54.1 (-1.5) && 35.6 (-1.9) & 18.3 (-1.0) && 51.2 (-2.2) & 55.4 (-2.1) && 67.5& 
98.9& 96.3& 76.1& 84.7 (-1.1)\\
w/o semantic &&  43.2 (-11.6)& 43.5 (-12.1) && 24.7 (-12.8) & 12.3 (-7.0) && 35.1 (-18.3) & 33.5 (-24.0) && 63.4& 94.5& 89.1& 65.8& 78.2 (-7.6)\\
w/o object && 51.3 (-3.5) & 51.4 (-4.2) && 34.1 (-3.4) & 16.9 (-2.4) && 50.2 (-3.2) & 54.1 (-3.4) && 65.2& 96.8& 94.9& 73.8& 82.7 (-3.1)\\
w/o box && 53.1 (-1.7) & 53.6 (-2.0) && 35.2 (-2.3) & 17.7 (-1.6) && 50.7 (-2.7)& 54.6 (-2.9) && 65.8& 97.1& 95.2& 75.3& 83.4 (-2.4)\\
w/o depth && 53.4 (-1.4) & 53.7 (-1.9) && 35.4 (-2.1) & 18.0 (-1.3) && 50.9 (-2.5) & 54.9 (-2.6) && 66.7& 98.2& 95.5& 75.7& 84.0 (-1.8)\\
% \rowcolor{gray!25} with all && 54.8 & 55.6 && 37.5 & 19.3 && 53.4 & 57.5 && 68.8& 99.6& 97.4& 77.4& 85.8\\
\bottomrule
\end{tabular}
}
\caption{Ablation study results in which scores from each evaluation level are excluded. The numbers in brackets indicate how much the method has decreased relative to the baseline. These results are assessed using Kendall's correlation coefficients $\tau_b$ and $\tau_c$ on the Flickr8k-Expert, Flickr8k-CF \cite{hodosh2013framing}, and Composite datasets \cite{aditya2015images}, along with pairwise ranking accuracy on the Pascal-50S dataset \cite{vedantam2015cider}. }
\label{tab:ablation}
\vspace{-0.35cm}
\end{table*}

\subsection{Implementation details}
In our framework, we utilize FLUX.1 [dev] \cite{flux_dev} as the text-to-image model. 
For the measurement of pixel level, we adopt the Euclidean distance metric, corresponding to p = 2 in \Cref{equ:pix}. This choice effectively quantifies the spatial discrepancies between corresponding pixel values. For the image feature extraction module, we employ the pre-trained ViT-B16 architecture, utilizing the image encoder component from CLIP \cite{radford2021learning}. 
For the object level, we utilize YOLOv8 \cite{yolov8} as the object detector due to its stable and excellent performance and efficiency in object detection tasks. Besides, we integrate Depth Anything V2 \cite{depth_anything_v2} to incorporate depth estimation capabilities, thereby enhancing the model's model's ability to analyze the 3D-spatial relationship within the scene. This combination of advanced object detection and depth estimation frameworks facilitates a more comprehensive analysis, ensuring robust performance across diverse application scenarios. Additional implementation details can be found in the Appendix.
% \textbf{With the training splits of Flickr30k \cite{young2014image} and MSCOCO \cite{lin2014microsoft} datasets,} training dataset? $\delta$ sigmoid function , where $\tau$ is the to-be-tuned temperature.
\subsection{Correlation with human judgment}
We first evaluate our CAMScore on the Flicker8k-Expert and Flicker8k-CF \cite{hodosh2013framing} datasets. Consistent with previous works, we compute the Kendall correlation coefficient $\tau_b$ and $\tau_c$. The results, as shown in \Cref{tab:flickr}, compare CAMScore's performance against the baseline metrics mentioned in \Cref{subsec:baselines}. The first part of the table is for reference-based methods, and the second part is for reference-free methods. The experimental results demonstrate that CAMScore exhibits a strong correlation with human judgments across considered datasets, thereby affirming its superiority over previously proposed metrics. In particular, CAMScore improves the Kendall correlation coefficient $\tau_b$ and $\tau_c$ by 0.9 and 1.3 points compared to PAC-S on the Flickr8k-Expert dataset, and by 1.5 and 0.7 points compared to PAC-S on the Flickr8k-CF dataset, respectively. Notably, CAMScore significantly outperforms the correlation coefficient achieved by evaluation metrics that necessitate reference captions,  highlighting its superior ability to align with human judgments without dependence on reference annotations.

We further conducted experiments on the Composite dataset \cite{aditya2015images}. The experiment results, shown in \Cref{tab:composite}, are also presented in terms of Kendall $\tau_b$ and Kendall $\tau_c$ correlation coefficient. CAMScore achieves the best correlation with human judgment among all baseline metrics, improving by 1.9 on the $\tau_b$ correlation coefficient and 1.8 on the $\tau_c$ correlation coefficient over the second highest score. These improvements demonstrate the comprehensive effectiveness of CAMScore in aligning with human judgments across diverse benchmarking scenarios.

\subsection{Accuracy on pairwise ranking}
We assess the effectiveness of CAMScore on the Pascal-50S dataset  \cite{vedantam2015cider}, which provides pairwise preference judgments between two captions. The caption that received the majority vote was deemed the preferred choice. In this setting, instead of computing Kendall correlation scores, we compute pairwise accuracy by considering for each pair the caption preferred by the majority of human ratings as correct and measuring how often the evaluation metric assigns a higher score to the selected caption. \Cref{tab:pascal} presents the accuracy results for Pascal-50S. In our experiments, CAMScore achieves state-of-the-art (SOTA) results with a mean accuracy of 85.8, proving the general effectiveness of our approach. Besides, CAMScore achieves state-of-the-art (SOTA) results with accuracies of 68.8 and 77.4 for HC and MM respectively. Although CAMScore does not attain the highest accuracy for the HI and HM, its performance remains closely comparable to existing SOTA accuracy. These results provide strong evidence that our metric is on par with the strong baselines and can therefore serve as a potent automatic evaluation metric for image captioning.

\subsection{Ablation study}
\begin{figure*}[t]
  \centering
  \includegraphics[width=\linewidth]{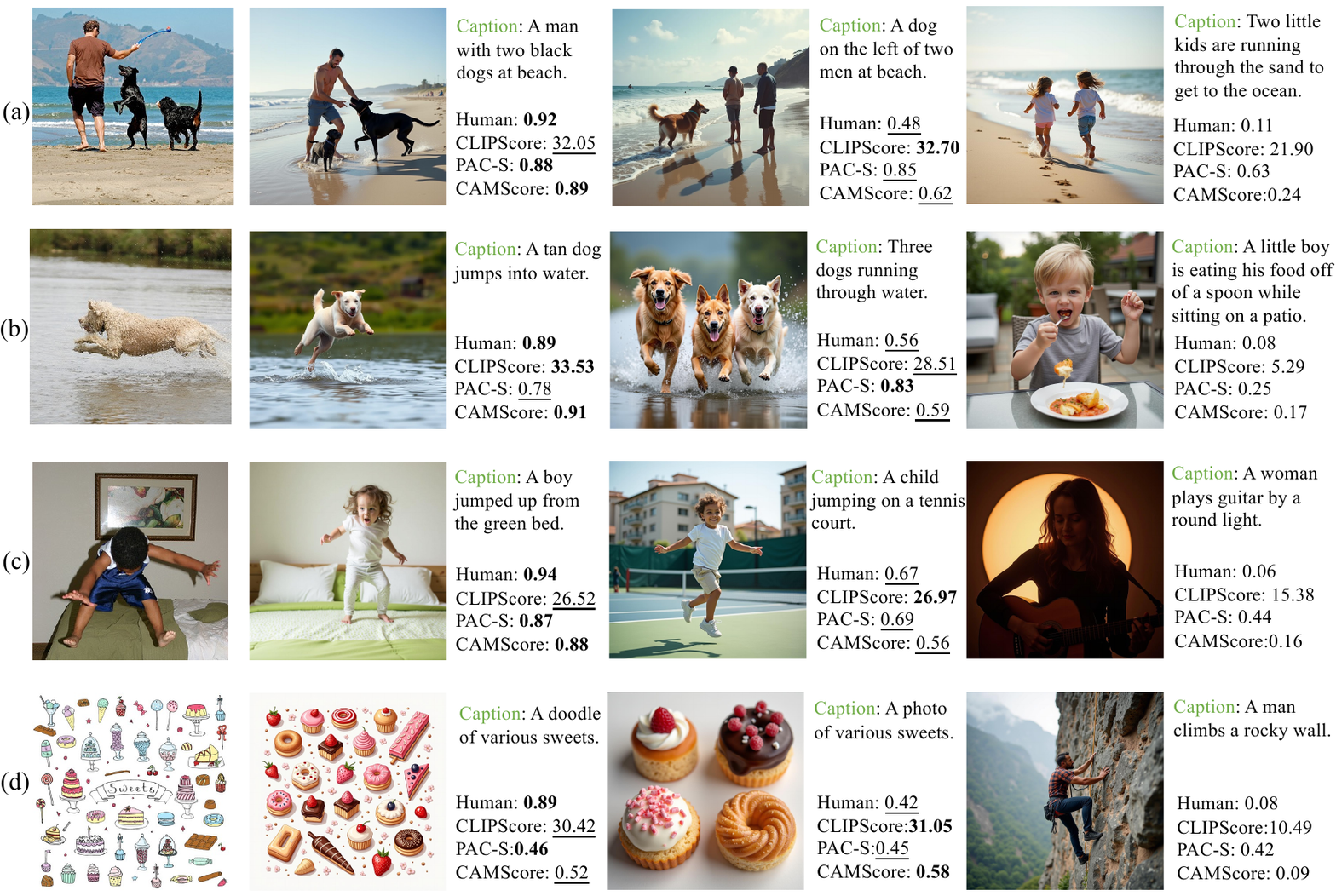}
  % \fbox{\rule{0pt}{4in} \rule{\linewidth}{0pt}}
   %\includegraphics[width=0.8\linewidth]{egfigure.eps}
   \caption{Examples of successful (a,b,c) and failed (d) cases. Except for the last non-photorealistic case, all others are from the Flickr8K dataset. The first column is the original image and the last three columns are the generated images with captions and metrics. Human judgment prefers the leftmost caption and dislikes the rightmost caption. Our metric is more consistent with human judgment. }
   \label{fig:case}
\end{figure*}
% Successful cases (a) numeracy case (b) position case (c) reference style case (d) Failure case on the non-photorealistic image.
We investigate the impact of various score integration strategies on overall performance. Specifically, we remove the scores of integration respectively to verify the impact of each score on the performance, as shown in \Cref{tab:ablation}. The performance without the semantic level metric is the lowest across all benchmark datasets, indicating that the semantic level poses an important role in the proposed metric. In addition, the impact of object level is the second largest, which illustrates the effectiveness of our design for object matching and quantity penalties. The absence of pixel level has the least impact on the performance, indicating that direct pixel differences are less relevant to human judgment.
By systematically evaluating the influence of each evaluated score, we verified that metrics at each level are necessary.

\subsection{Case study}
\Cref{fig:case} shows various example of CAMScore. For each case, we provide three different captions along with their corresponding scores. All methods are capable of distinguishing irrelevant captions for photo-style images, as illustrated in the last column. However, performance diminishes when other metrics deal with more detailed problems. Specifically, CLIPScore confuses the positional relationship and corresponding numeracy of man and dogs, and confuses the scenes where the child is for the first two columns of \Cref{fig:case} (a) and (c), respectively. PAS-S confuses the numeracy of dogs for the first two columns of \Cref{fig:case} (b). In contrast, CAMScore accurately captures attribute and location information, aligning closely with human judgment.

\subsection{Limitations and future works}
CAMScore is constrained by the detection performance of the object detector employed in the framework. Current SOTA object detection models are primarily based on the MS-COCO dataset, which has a limited number of specific categories. Besides, object detectors generally exhibit suboptimal performance for non-photorealistic images, leading to failure of CAMScore, as shown in the last row in \Cref{fig:case}. These limitations may be eliminated with the future development of more powerful object detectors. In addition, text-to-image models usually require longer processing times and more computational resources compared to object detection models.
In future work, we aim to incorporate the metric into model training procedures to enhance specific capabilities.

% Besides, object detectors generally exhibit suboptimal performance for non-photorealistic images. These limitations may be eliminated with the future development of more powerful object detectors. In addition, the computational resources and inference time of the text-to-image model also need to be taken into consideration. Typically, these models demand longer processing times and more computational resources compared to object detection models. Typically, these models demand longer to explore the cycle consistency across other modalities and 
\section{Conclusion}
In this paper, we proposed CAMScore, a novel cyclic reference-free evaluation metric for image captioning. CAMScore enables reference-free evaluation while addressing the modality gap inherent in existing cross-modality metrics by leveraging the cycle consistency between image captioning and text-to-image generation. Additionally, our three-level modular framework provides fine-grained information for evaluation from the pixel-level, semantic-level, and object-level. Furthermore, experimental comparisons with baseline metrics across various benchmarks demonstrate that CAMScore has a strong correlation with human judgment, showing the effectiveness of the proposed metric.

\clearpage

{
    \small
    \bibliographystyle{ieeenat_fullname}
    \bibliography{main}

\begin{thebibliography}{47}
\providecommand{\natexlab}[1]{#1}
\providecommand{\url}[1]{\texttt{#1}}
\expandafter\ifx\csname urlstyle\endcsname\relax
  \providecommand{\doi}[1]{doi: #1}\else
  \providecommand{\doi}{doi: \begingroup \urlstyle{rm}\Url}\fi

\bibitem[Aditya et~al.(2015)Aditya, Yang, Baral, Fermuller, and Aloimonos]{aditya2015images}
Somak Aditya, Yezhou Yang, Chitta Baral, Cornelia Fermuller, and Yiannis Aloimonos.
\newblock {From Images to Sentences through Scene Description Graphs using Commonsense Reasoning and Knowledge}.
\newblock \emph{arXiv preprint arXiv:1511.03292}, 2015.

\bibitem[Anderson et~al.(2016)Anderson, Fernando, Johnson, and Gould]{anderson2016spice}
Peter Anderson, Basura Fernando, Mark Johnson, and Stephen Gould.
\newblock {SPICE}: Semantic propositional image caption evaluation.
\newblock In \emph{ECCV}, pages 382--398, 2016.

\bibitem[Anderson et~al.(2018)Anderson, He, Buehler, Teney, Johnson, Gould, and Zhang]{anderson2018bottom}
Peter Anderson, Xiaodong He, Chris Buehler, Damien Teney, Mark Johnson, Stephen Gould, and Lei Zhang.
\newblock {Bottom-Up and Top-Down Attention for Image Captioning and Visual Question Answering}.
\newblock In \emph{CVPR}, pages 6077--6086, 2018.

\bibitem[Bai et~al.(2022)Bai, Liu, Ni, Wang, Hu, Guo, and Cheng]{bai2022lat}
Jinbin Bai, Chunhui Liu, Feiyue Ni, Haofan Wang, Mengying Hu, Xiaofeng Guo, and Lele Cheng.
\newblock {LaT: Latent Translation with Cycle-Consistency for Video-Text Retrieval}.
\newblock \emph{arXiv preprint arXiv:2207.04858}, 2022.

\bibitem[Bai et~al.(2024)Bai, Ye, Chow, Song, Chen, Li, Dong, Zhu, and Yan]{bai2024meissonic}
Jinbin Bai, Tian Ye, Wei Chow, Enxin Song, Qing-Guo Chen, Xiangtai Li, Zhen Dong, Lei Zhu, and Shuicheng Yan.
\newblock {Meissonic: Revitalizing Masked Generative Transformers for Efficient High-Resolution Text-to-Image Synthesis}.
\newblock \emph{arXiv preprint arXiv:2410.08261}, 2024.

\bibitem[Banerjee and Lavie(2005)]{banerjee2005meteor}
Satanjeev Banerjee and Alon Lavie.
\newblock {METEOR: An Automatic Metric for MT Evaluation with Improved Correlation with Human Judgments}.
\newblock In \emph{Proceedings of the {ACL} Workshop on Intrinsic and Extrinsic Evaluation Measures for Machine Translation and/or Summarization}, pages 65--72. Association for Computational Linguistics, 2005.

\bibitem[Chen et~al.(2020)Chen, Li, Yu, El~Kholy, Ahmed, Gan, Cheng, and Liu]{chen2020uniter}
Yen-Chun Chen, Linjie Li, Licheng Yu, Ahmed El~Kholy, Faisal Ahmed, Zhe Gan, Yu Cheng, and Jingjing Liu.
\newblock {UNITER: UNiversal Image-TExt Representation Learning}.
\newblock In \emph{ECCV}, pages 104--120, 2020.

\bibitem[Devlin et~al.(2019)Devlin, Chang, Lee, and Toutanova]{devlin2019bert}
Jacob Devlin, Ming-Wei Chang, Kenton Lee, and Kristina Toutanova.
\newblock {BERT: Pre-training of Deep Bidirectional Transformers for Language Understanding}.
\newblock In \emph{Proceedings of the 2019 Conference of the North American Chapter of the Association for Computational Linguistics: Human Language Technologies, Volume 1 (Long and Short Papers)}, pages 4171--4186, 2019.

\bibitem[Eigen et~al.(2014)Eigen, Puhrsch, and Fergus]{eigen2014depth}
David Eigen, Christian Puhrsch, and Rob Fergus.
\newblock {Depth Map Prediction from a Single Image using a Multi-Scale Deep Network}.
\newblock In \emph{NeurIPS}, pages 2366--2374, 2014.

\bibitem[Godard et~al.(2017)Godard, Mac~Aodha, and Brostow]{godard2017unsupervised}
Cl{\'e}ment Godard, Oisin Mac~Aodha, and Gabriel~J Brostow.
\newblock {Unsupervised Monocular Depth Estimation with Left-Right Consistency}.
\newblock In \emph{CVPR}, pages 270--279, 2017.

\bibitem[Guo et~al.(2019)Guo, Liu, Yao, Li, and Lu]{guo2019mscap}
Longteng Guo, Jing Liu, Peng Yao, Jiangwei Li, and Hanqing Lu.
\newblock {MSCap: Multi-Style Image Captioning with Unpaired Stylized Text}.
\newblock In \emph{CVPR}, pages 4204--4213, 2019.

\bibitem[Gurari et~al.(2020)Gurari, Zhao, Zhang, and Bhattacharya]{gurari2020captioning}
Danna Gurari, Yinan Zhao, Meng Zhang, and Nilavra Bhattacharya.
\newblock {Captioning Images Taken by People Who Are Blind}.
\newblock In \emph{ECCV}, pages 417--434. Springer, 2020.

\bibitem[Hessel et~al.(2021)Hessel, Holtzman, Forbes, Le~Bras, and Choi]{hessel2021clipscore}
Jack Hessel, Ari Holtzman, Maxwell Forbes, Ronan Le~Bras, and Yejin Choi.
\newblock {CLIPScore: A Reference-free Evaluation Metric for Image Captioning}.
\newblock In \emph{Proceedings of the 2021 Conference on Empirical Methods in Natural Language Processing}, pages 7514--7528. Association for Computational Linguistics, 2021.

\bibitem[Hodosh et~al.(2013)Hodosh, Young, and Hockenmaier]{hodosh2013framing}
Micah Hodosh, Peter Young, and Julia Hockenmaier.
\newblock Framing image description as a ranking task: {D}ata, models and evaluation metrics.
\newblock \emph{Journal of Artificial Intelligence Research}, 47:\penalty0 853--899, 2013.

\bibitem[Hoffman et~al.(2018)Hoffman, Tzeng, Park, Zhu, Isola, Saenko, Efros, and Darrell]{hoffman2018cycada}
Judy Hoffman, Eric Tzeng, Taesung Park, Jun-Yan Zhu, Phillip Isola, Kate Saenko, Alexei Efros, and Trevor Darrell.
\newblock {CyCADA: Cycle-Consistent Adversarial Domain Adaptation}.
\newblock In \emph{ICML}, pages 1989--1998. Pmlr, 2018.

\bibitem[Hu et~al.(2021)Hu, Chen, and Jin]{hu2021question}
Anwen Hu, Shizhe Chen, and Qin Jin.
\newblock {Question-controlled Text-aware Image Captioning}.
\newblock In \emph{ACM MM}, pages 3097--3105, 2021.

\bibitem[Hu et~al.(2023)Hu, Chen, Zhang, and Jin]{hu2023infometic}
Anwen Hu, Shizhe Chen, Liang Zhang, and Qin Jin.
\newblock {I}nfo{M}et{IC}: An informative metric for reference-free image caption evaluation.
\newblock In \emph{Proceedings of the 61st Annual Meeting of the Association for Computational Linguistics (Volume 1: Long Papers)}, pages 3171--3185. Association for Computational Linguistics, 2023.

\bibitem[Jiang et~al.(2019)Jiang, Huang, Zhang, Wang, Zhang, Gan, Diesner, and Gao]{jiang2019tiger}
Ming Jiang, Qiuyuan Huang, Lei Zhang, Xin Wang, Pengchuan Zhang, Zhe Gan, Jana Diesner, and Jianfeng Gao.
\newblock {TIGEr: Text-to-Image Grounding for Image Caption Evaluation}.
\newblock In \emph{Proceedings of the 2019 Conference on Empirical Methods in Natural Language Processing and the 9th International Joint Conference on Natural Language Processing (EMNLP-IJCNLP)}, pages 2141--2152. Association for Computational Linguistics, 2019.

\bibitem[Jocher et~al.(2023)Jocher, Chaurasia, and Qiu]{yolov8}
Glenn Jocher, Ayush Chaurasia, and Jing Qiu.
\newblock {Ultralytics YOLOv8}, 2023.

\bibitem[Kim et~al.(2022)Kim, Kim, Lee, Yoo, and Lee]{kim2022mutual}
Jin-Hwa Kim, Yunji Kim, Jiyoung Lee, Kang~Min Yoo, and Sang-Woo Lee.
\newblock {Mutual Information Divergence: A Unified Metric for Multimodal Generative Models}.
\newblock In \emph{NeurIPS}, pages 35072--35086, 2022.

\bibitem[Kuhn(1955)]{kuhn1955hungarian}
Harold~W Kuhn.
\newblock {The Hungarian Method for the Assignment Problem}.
\newblock \emph{Naval research logistics quarterly}, 2\penalty0 (1-2):\penalty0 83--97, 1955.

\bibitem[Labs(2024)]{flux_dev}
Black~Forest Labs.
\newblock {FLUX.1 [dev]}, 2024.

\bibitem[Lee et~al.(2020)Lee, Yoon, Dernoncourt, Kim, Bui, and Jung]{lee2020vilbertscore}
Hwanhee Lee, Seunghyun Yoon, Franck Dernoncourt, Doo~Soon Kim, Trung Bui, and Kyomin Jung.
\newblock {V}i{LBERTS}core: Evaluating image caption using vision-and-language {BERT}.
\newblock In \emph{Proceedings of the First Workshop on Evaluation and Comparison of NLP Systems}, pages 34--39. Association for Computational Linguistics, 2020.

\bibitem[Lee et~al.(2021)Lee, Yoon, Dernoncourt, Bui, and Jung]{lee2021umic}
Hwanhee Lee, Seunghyun Yoon, Franck Dernoncourt, Trung Bui, and Kyomin Jung.
\newblock {UMIC: An Unreferenced Metric for Image Captioning via Contrastive Learning}.
\newblock In \emph{Proceedings of the 59th Annual Meeting of the Association for Computational Linguistics and the 11th International Joint Conference on Natural Language Processing (Volume 2: Short Papers)}, pages 220--226, 2021.

\bibitem[Lee et~al.(2024)Lee, Park, and Kang]{lee2024fleur}
Yebin Lee, Imseong Park, and Myungjoo Kang.
\newblock {FLEUR: An Explainable Reference-Free Evaluation Metric for Image Captioning Using a Large Multimodal Model}.
\newblock In \emph{Proceedings of the 62nd Annual Meeting of the Association for Computational Linguistics (Volume 1: Long Papers)}, pages 3732--3746. Association for Computational Linguistics, 2024.

\bibitem[Liang et~al.(2022)Liang, Zhang, Kwon, Yeung, and Zou]{liang2022mind}
Victor~Weixin Liang, Yuhui Zhang, Yongchan Kwon, Serena Yeung, and James~Y Zou.
\newblock {Mind the Gap: Understanding the Modality Gap in Multi-modal Contrastive Representation Learning}.
\newblock In \emph{NeurIPS}, pages 17612--17625, 2022.

\bibitem[Lin(2004)]{lin2004rouge}
Chin-Yew Lin.
\newblock {ROUGE}: A package for automatic evaluation of summaries.
\newblock In \emph{Text summarization branches out}, pages 74--81, 2004.

\bibitem[Lin et~al.(2014)Lin, Maire, Belongie, Hays, Perona, Ramanan, Doll{\'a}r, and Zitnick]{lin2014microsoft}
Tsung-Yi Lin, Michael Maire, Serge Belongie, James Hays, Pietro Perona, Deva Ramanan, Piotr Doll{\'a}r, and C~Lawrence Zitnick.
\newblock {Microsoft COCO: Common Objects in Context}.
\newblock In \emph{ECCV}, pages 740--755. Springer, 2014.

\bibitem[Lu et~al.(2019)Lu, Batra, Parikh, and Lee]{lu2019vilbert}
Jiasen Lu, Dhruv Batra, Devi Parikh, and Stefan Lee.
\newblock {ViLBERT: Pretraining Task-Agnostic Visiolinguistic Representations for Vision-and-Language Tasks}.
\newblock In \emph{NeurIPS}, pages 13--23, 2019.

\bibitem[Munkres(1957)]{munkres1957algorithms}
James Munkres.
\newblock {Algorithms for the Assignment and Transportation Problems}.
\newblock \emph{Journal of the society for industrial and applied mathematics}, 5\penalty0 (1):\penalty0 32--38, 1957.

\bibitem[Papineni et~al.(2002)Papineni, Roukos, Ward, and Zhu]{papineni2002bleu}
Kishore Papineni, Salim Roukos, Todd Ward, and Wei-Jing Zhu.
\newblock {BLEU}: a method for automatic evaluation of machine translation.
\newblock In \emph{Proceedings of the 40th annual meeting of the Association for Computational Linguistics}, pages 311--318, 2002.

\bibitem[Radford et~al.(2021)Radford, Kim, Hallacy, Ramesh, Goh, Agarwal, Sastry, Askell, Mishkin, Clark, et~al.]{radford2021learning}
Alec Radford, Jong~Wook Kim, Chris Hallacy, Aditya Ramesh, Gabriel Goh, Sandhini Agarwal, Girish Sastry, Amanda Askell, Pamela Mishkin, Jack Clark, et~al.
\newblock {Learning Transferable Visual Models From Natural Language Supervision}.
\newblock In \emph{International conference on machine learning}, pages 8748--8763. PMLR, 2021.

\bibitem[Sarto et~al.(2023)Sarto, Barraco, Cornia, Baraldi, and Cucchiara]{sarto2023positive}
Sara Sarto, Manuele Barraco, Marcella Cornia, Lorenzo Baraldi, and Rita Cucchiara.
\newblock {Positive-Augmented Contrastive Learning for Image and Video Captioning Evaluation}.
\newblock In \emph{CVPR}, pages 6914--6924, 2023.

\bibitem[Shah et~al.(2019)Shah, Chen, Rohrbach, and Parikh]{shah2019cycle}
Meet Shah, Xinlei Chen, Marcus Rohrbach, and Devi Parikh.
\newblock {Cycle-Consistency for Robust Visual Question Answering}.
\newblock In \emph{CVPR}, pages 6649--6658, 2019.

\bibitem[Shi et~al.(2023)Shi, Welle, Bj{\"o}rkman, and Kragic]{shi2023towards}
Peiyang Shi, Michael~C. Welle, M{\r{a}}rten Bj{\"o}rkman, and Danica Kragic.
\newblock Towards understanding the modality gap in {CLIP}.
\newblock In \emph{ICLR 2023 Workshop on Multimodal Representation Learning: Perks and Pitfalls}, 2023.

\bibitem[Vedantam et~al.(2015)Vedantam, Lawrence~Zitnick, and Parikh]{vedantam2015cider}
Ramakrishna Vedantam, C Lawrence~Zitnick, and Devi Parikh.
\newblock {CIDE}r: Consensus-based image description evaluation.
\newblock In \emph{CVPR}, pages 4566--4575, 2015.

\bibitem[Wada et~al.(2024)Wada, Kaneda, Saito, and Sugiura]{wada2024polos}
Yuiga Wada, Kanta Kaneda, Daichi Saito, and Komei Sugiura.
\newblock {Polos: Multimodal Metric Learning from Human Feedback for Image Captioning}.
\newblock In \emph{CVPR}, pages 13559--13568, 2024.

\bibitem[Wang et~al.(2019)Wang, Song, Ma, Zhou, Liu, and Li]{wang2019unsupervised}
Ning Wang, Yibing Song, Chao Ma, Wengang Zhou, Wei Liu, and Houqiang Li.
\newblock {Unsupervised Deep Tracking}.
\newblock In \emph{CVPR}, pages 1308--1317, 2019.

\bibitem[Wang et~al.(2021)Wang, Yao, Wang, Wu, and Chen]{wang2021faier}
Sijin Wang, Ziwei Yao, Ruiping Wang, Zhongqin Wu, and Xilin Chen.
\newblock {FAIEr: Fidelity and Adequacy Ensured Image Caption Evaluation}.
\newblock In \emph{CVPR}, pages 14050--14059, 2021.

\bibitem[Xu et~al.(2015)Xu, Ba, Kiros, Cho, Courville, Salakhudinov, Zemel, and Bengio]{xu2015show}
Kelvin Xu, Jimmy Ba, Ryan Kiros, Kyunghyun Cho, Aaron Courville, Ruslan Salakhudinov, Rich Zemel, and Yoshua Bengio.
\newblock Show, attend and tell: Neural image caption generation with visual attention.
\newblock In \emph{ICML}, pages 2048--2057. PMLR, 2015.

\bibitem[Yang et~al.(2024)Yang, Kang, Huang, Zhao, Xu, Feng, and Zhao]{depth_anything_v2}
Lihe Yang, Bingyi Kang, Zilong Huang, Zhen Zhao, Xiaogang Xu, Jiashi Feng, and Hengshuang Zhao.
\newblock {Depth Anything V2}.
\newblock \emph{arXiv preprint arXiv:2406.09414}, 2024.

\bibitem[Yi et~al.(2020)Yi, Deng, and Hu]{yi2020improving}
Yanzhi Yi, Hangyu Deng, and Jinglu Hu.
\newblock {Improving Image Captioning Evaluation by Considering Inter References Variance}.
\newblock In \emph{Proceedings of the 58th Annual Meeting of the Association for Computational Linguistics}, pages 985--994, 2020.

\bibitem[Young et~al.(2014)Young, Lai, Hodosh, and Hockenmaier]{young2014image}
Peter Young, Alice Lai, Micah Hodosh, and Julia Hockenmaier.
\newblock {From image descriptions to visual denotations: New similarity metrics for semantic inference over event descriptions}.
\newblock \emph{Transactions of the Association for Computational Linguistics}, 2:\penalty0 67--78, 2014.

\bibitem[Zhang et~al.(2019)Zhang, Kishore, Wu, Weinberger, and Artzi]{zhang2019bertscore}
Tianyi Zhang, Varsha Kishore, Felix Wu, Kilian~Q Weinberger, and Yoav Artzi.
\newblock {BERTS}core: {E}valuating {T}ext {G}eneration with {BERT}.
\newblock In \emph{ICLR}, 2019.

\bibitem[Zhao et~al.(2019)Zhao, Peyrard, Liu, Gao, Meyer, and Eger]{zhao2019moverscore}
Wei Zhao, Maxime Peyrard, Fei Liu, Yang Gao, Christian~M. Meyer, and Steffen Eger.
\newblock {M}over{S}core: {T}ext {G}eneration {E}valuating with {C}ontextualized {E}mbeddings and {E}arth {M}over {D}istance.
\newblock In \emph{Proceedings of the 2019 Conference on Empirical Methods in Natural Language Processing and the 9th International Joint Conference on Natural Language Processing (EMNLP-IJCNLP)}, pages 563--578. Association for Computational Linguistics, 2019.

\bibitem[Zheng et~al.(2021)Zheng, Wang, Ren, Liu, Ye, Hu, and Zuo]{zheng2021enhancing}
Zhaohui Zheng, Ping Wang, Dongwei Ren, Wei Liu, Rongguang Ye, Qinghua Hu, and Wangmeng Zuo.
\newblock {Enhancing Geometric Factors in Model Learning and Inference for Object Detection and Instance Segmentation}.
\newblock \emph{IEEE transactions on cybernetics}, 52\penalty0 (8):\penalty0 8574--8586, 2021.

\bibitem[Zhu et~al.(2017)Zhu, Park, Isola, and Efros]{zhu2017unpaired}
Jun-Yan Zhu, Taesung Park, Phillip Isola, and Alexei~A. Efros.
\newblock {Unpaired Image-to-Image Translation Using Cycle-Consistent Adversarial Networks}.
\newblock In \emph{ICCV}, pages 2242--2251, 2017.

\end{thebibliography}
}

\clearpage
\setcounter{page}{1}
\maketitlesupplementary

\begin{figure*}[hb]
  \centering
  \includegraphics[width=\linewidth]{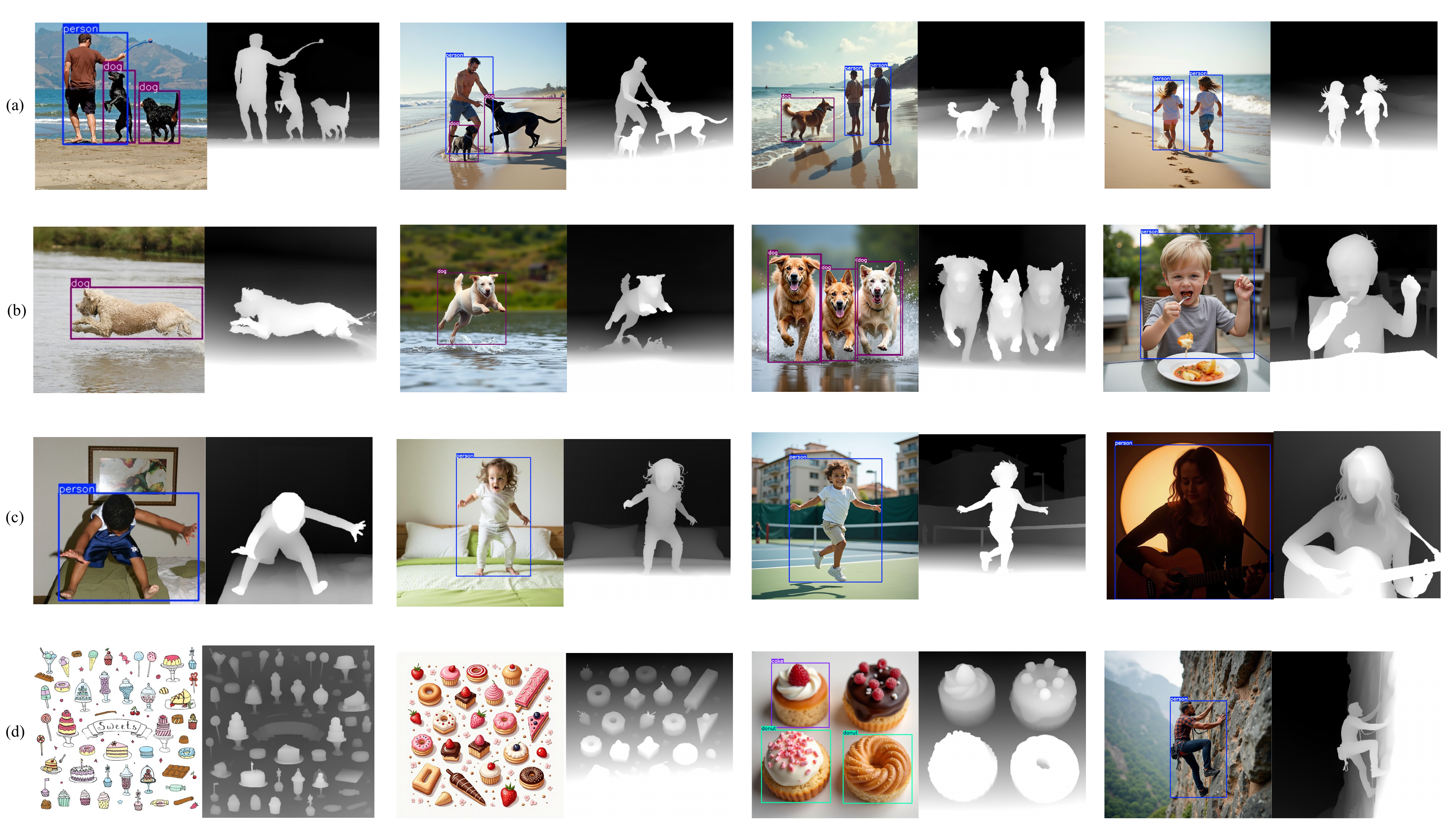}
  % \fbox{\rule{0pt}{4in} \rule{\linewidth}{0pt}}
   %\includegraphics[width=0.8\linewidth]{egfigure.eps}
   \caption{Object detection and depth estimation result of case study. The object detector successfully boxes the object in case (a,b,c), while failing in the non-photorealistic case (d).}
   \label{fig:app}
\end{figure*}
\section{Implementation details}
\subsection{Dataset Setup}

\tinytit{Flickr8k-Expert} \cite{hodosh2013framing} Flickr8k-Expert dataset contains 17k expert annotations for image-caption pairs, encompassing a total of 5,664 distinct images. The captions in Flickr 8K were selected by an image retrieval system from a reference caption pool, instead of generated using a learned image captioning model. Each image-caption pair is evaluated by three expert annotators with scores ranging from 1 to 4, where a score of 1 indicates that the caption does not correlate with the image and a score of 4 signifies that the caption describes the corresponding image without errors. This evaluation methodology ensures a reliable assessment of caption quality in relation to the visual content.\\

\tinytit{Flickr8k-CF} \cite{hodosh2013framing} Flickr8k-CF consists of 145k binary quality judgments, collected from CrowdFlower, covering 48k image-caption pairs that contain 1k unique images. For each pair, three annotators determine whether the image-caption pair is a valid match, categorizing their responses as either “yes” or “no”. The final score for each image-caption pair is calculated as the proportion of “yes” responses. \\

\tinytit{Composite} \cite{aditya2015images} Composite dataset comprises human judgments of 12,000 image-caption pairs, incorporating 3,995 images from MSCOCO (2,007 images) \cite{lin2014microsoft}, Flickr30K (991 images) \cite{young2014image}, and Flickr8k (997 images) \cite{hodosh2013framing}. Candidate captions were sourced from both human reference captions and two captioning models. In this dataset, human evaluators score each caption's relevance to its corresponding image using a five-point scale, where 1 indicates that “the description has nothing to do with the image” and 5 signifies that “the description is perfectly relevant to the image.” \\

\tinytit{Pascal-50S} \cite{vedantam2015cider} Pascal-50S includes 4,000 caption pairs associated with 1,000 images from UIUC PASCAL Sentence Dataset. Each pair is annotated to indicate which of the two captions is considered correct by 48 annotators. Each image is linked to approximately 50 reference captions generated by humans. Annotators are tasked with selecting the candidate caption in each pair that most closely aligns with the provided reference descriptions. In Pascal-50S, caption pairs are categorized based on the composition of the two captions: 1) Human-Human Correct (HC) contains two human-written captions for the target image, 2) Human-Human Incorrect (HI) includes two captions written by humans but describing different images, 3) the group of Human-Machine (HM) contains a human written and a machine-generated caption, and 4) Machine-Machine (MM) includes two matching generated captions focusing on the same image.

\subsection{Experiment details}
In our framework, we utilize FLUX.1 [dev] \cite{flux_dev} as the text-to-image model. 
For the measurement of pixel level, we adopt the Euclidean distance metric, corresponding to p = 2. This choice effectively quantifies the spatial discrepancies between corresponding pixel values. For the image feature extraction module, we employ the pre-trained ViT-B16 architecture, utilizing the image encoder component from CLIP \cite{radford2021learning}. In the ViT-B16 architecture, the input image is divided into patches, which are then transformed into vectors through embedding layers. These vectors are subsequently processed by Transformer blocks that consist of self-attention and feed-forward layers.
For the object level, we utilize YOLOv8x, the largest model in the YOLOv8 family as the object detector due to its stable and excellent performance and efficiency in object detection tasks. Besides, we integrate Depth Anything v2-Large, which is also the largest model currently available in the Depth Anything v2 family to incorporate depth estimation capabilities, thereby enhancing the model's model's ability to analyze the 3D-spatial relationship within the scene. This combination of advanced object detection and depth estimation frameworks facilitates a more comprehensive analysis, ensuring robust performance across diverse application scenarios. When evaluating the original image with the generated image, we resize the original image to $512 \times 512$ for ease of calculation. The inputs are first transformed by the sigmoid function and then passed to the multilayer perception.  We use a batch size of 64, a learning rate of 3e-5, and used early stopping in our model to optimize for the highest Kendall’s $\tau$. We implement CAMScore with PyTorch and train on a NVIDIA A100 GPU. We calculate the Kendall tau score using the SciPy 1.14.1 implementation. 
\section{Additional Results on case study}
We provide the results of object detection and depth estimation for each case in the case study. As shown in \Cref{fig:app}, the object detector successfully boxes the objects in case
(a,b,c), while failing in the non-photorealistic case (d), and the failure of object detectors will affect our metric. The depth estimation performs well on objects across all cases, indicating that the depth estimation method to estimate the depth of image objects is robust. Since our object level evaluation mainly relies on object-matching relations, the performance of object detectors will affect our metric. The failure of the object detector on the non-photorealistic image \Cref{fig:app} (d) is the reason why this case fails, indicating that the current object detector has a lot of room for improvement.

\end{document}